\documentclass[runningheads]{llncs}
\usepackage{graphicx}
\usepackage{amsmath,amssymb} 
\usepackage{color}
\usepackage{multirow}
\usepackage{array}
\usepackage{xspace}
\usepackage{pifont}
\usepackage[width=122mm,left=12mm,paperwidth=146mm,height=193mm,top=12mm,paperheight=217mm]{geometry}
\usepackage{subfig}

\newcommand{\mat}[1]{\mathtt{#1}}
\newcommand{\set}[1]{\ensuremath{\mathcal{#1}}}
\newcommand{\con}[1]{\ensuremath{\mathsf{#1}}}

\newcommand{\myparagraph}[1]{\vspace{5pt}\noindent\textbf{#1}}

\DeclareMathOperator\sign{sgn}
\makeatletter
\DeclareRobustCommand\onedot{\futurelet\@let@token\@onedot}
\def\@onedot{\ifx\@let@token.\else.\null\fi\xspace}

\def\eg{\emph{e.g}\onedot} 
\def\ie{\emph{i.e}\onedot}

\newlength\eqcol@newlen
\newlength\eqcol@oldlen
\let\eqcol@bc\hfil
\let\eqcol@ec\hfil
\let\eqcol@br\hfil
\let\eqcol@el\hfil
\newcolumntype{e}[1]{%
  >{\setbox0\hbox\bgroup}#1%
  <{\egroup
    \ifdim\wd0<\eqcol@newlen\else\global\eqcol@newlen\wd0\fi
    \ifdim\wd0<\eqcol@oldlen\else\global\eqcol@oldlen\wd0\fi
    \hbox to \eqcol@oldlen{%
      \csname eqcol@b#1\endcsname
      \box0 %
      \csname eqcol@e#1\endcsname
    }%
  }%
}
\newcount\eqcol@count
\def\eqcolRead{%
  \global\advance\eqcol@count1 %
  \eqcol@oldlen5em\relax
  \csname eqcol@def@\romannumeral\eqcol@count\endcsname
}
\def\eqcolWrite{%
  \immediate\write\@auxout{%
  \gdef\expandafter\noexpand\csname eqcol@def@\romannumeral\eqcol@count\endcsname
    {\global\eqcol@oldlen\the\eqcol@newlen\relax}%
  }%
  \global\eqcol@newlen0pt\relax
}
\let\eqcol@old@tabular\tabular
\def\tabular{\eqcolRead\eqcol@old@tabular}
\let\eqcol@old@endtabular\endtabular
\def\endtabular{\eqcol@old@endtabular\eqcolWrite}
\makeatother

\begin{document}
\pagestyle{headings}
\mainmatter

\title{Adding New Tasks to a Single Network with Weight Transformations using Binary Masks}
\titlerunning{ }

\authorrunning{ }
\author{Massimiliano Mancini$^{1,2}$, Elisa Ricci$^{2,3}$, Barbara Caputo$^{1,4}$, Samuel Rota Bul\`o$^{5}$}
\institute{$^1$Sapienza University of Rome, $^2$Fondazione Bruno Kessler,$^3$University of Trento, $^4$Italian Institute of Technology, $^5$Mapillary Research\\
	\email{ \{mancini,caputo\}@diag.uniroma1.it,eliricci@fbk.eu,samuel@mapillary.com}
}

\maketitle

\begin{abstract}

Visual recognition algorithms are required today to exhibit adaptive abilities. Given a deep model trained on a 
specific, given task, it would be highly desirable to be able to adapt incrementally to new tasks, 
preserving scalability as the number of new tasks increases, while at the same time avoiding catastrophic forgetting issues. 
Recent work has shown that masking the internal weights of a given original conv-net through learned binary variables is a promising strategy.
We build upon this intuition and take into account more elaborated affine transformations of the convolutional weights that include learned binary masks.
We show that with our generalization it is possible to achieve significantly higher levels of adaptation to new tasks, enabling the approach to compete with fine tuning strategies by requiring slightly more than 1 bit per network parameter per additional task. 
Experiments on two popular benchmarks showcase the power of our approach, that achieves the new state of the art on the Visual Decathlon Challenge. 
\end{abstract}

\section{Introduction}\label{sec:intro}
A long-standing goal of Artificial Intelligence is the ability to adapt an initial, pre-trained model to 
novel, unseen scenarios. 
This is crucial for increasing the knowledge of an intelligent system and developing effective incremental \cite{ring1997child,KuzborskijOC13}, life-long learning \cite{thrun1995lifelong,thrun2012learning,silver2013lifelong} algorithms. While fascinating, achieving this goal requires facing multiple challenges. First, 
learning a new task should not negatively affect the performance on old tasks. Second, it should be avoided adding multiple parameters to the model for each new task learned, as it would lead to poor scalability of the framework. 
In this context, while deep learning algorithms have achieved impressive results on many computer vision benchmarks \cite{krizhevsky2012imagenet,he2016deep,girshick2014rich,long2015fully}, mainstream approaches for adapting deep models to novel tasks tend to suffer from the  problems  mentioned above. In fact, fine-tuning a given architecture to new data does produce a powerful model on the novel task, at the expense of a degraded performance on the old ones, resulting in the well-known phenomenon  of  catastrophic forgetting \cite{french1999catastrophic,goodfellow2013empirical}. At the same time, replicating the network parameters and training a separate network for each task is a powerful approach that preserves performances on old tasks, but at the cost of an explosion of the network parameters \cite{rebuffi2017learning}.

Different works addressed these problems by either considering losses encouraging the preservation of the current weights \cite{li2017learning,kirkpatrick2017overcoming} 
or by designing task-specific network parameters \cite{rusu2016progressive,rebuffi2017learning,rosenfeld2017incremental,mallya2017packnet,mallya2018piggyback}. Interestingly, in \cite{mallya2018piggyback} the authors showed that an effective strategy for achieving good incremental learning performances 
is to create a binary mask for each task. This binary mask is then multiplied by the main network weights, determining which of them are useful for addressing the new task. The total increase of network size required by their method is just one bit for each parameter per task. 

 \begin{figure}[t]
 \centering
 \includegraphics[width=0.8\columnwidth]{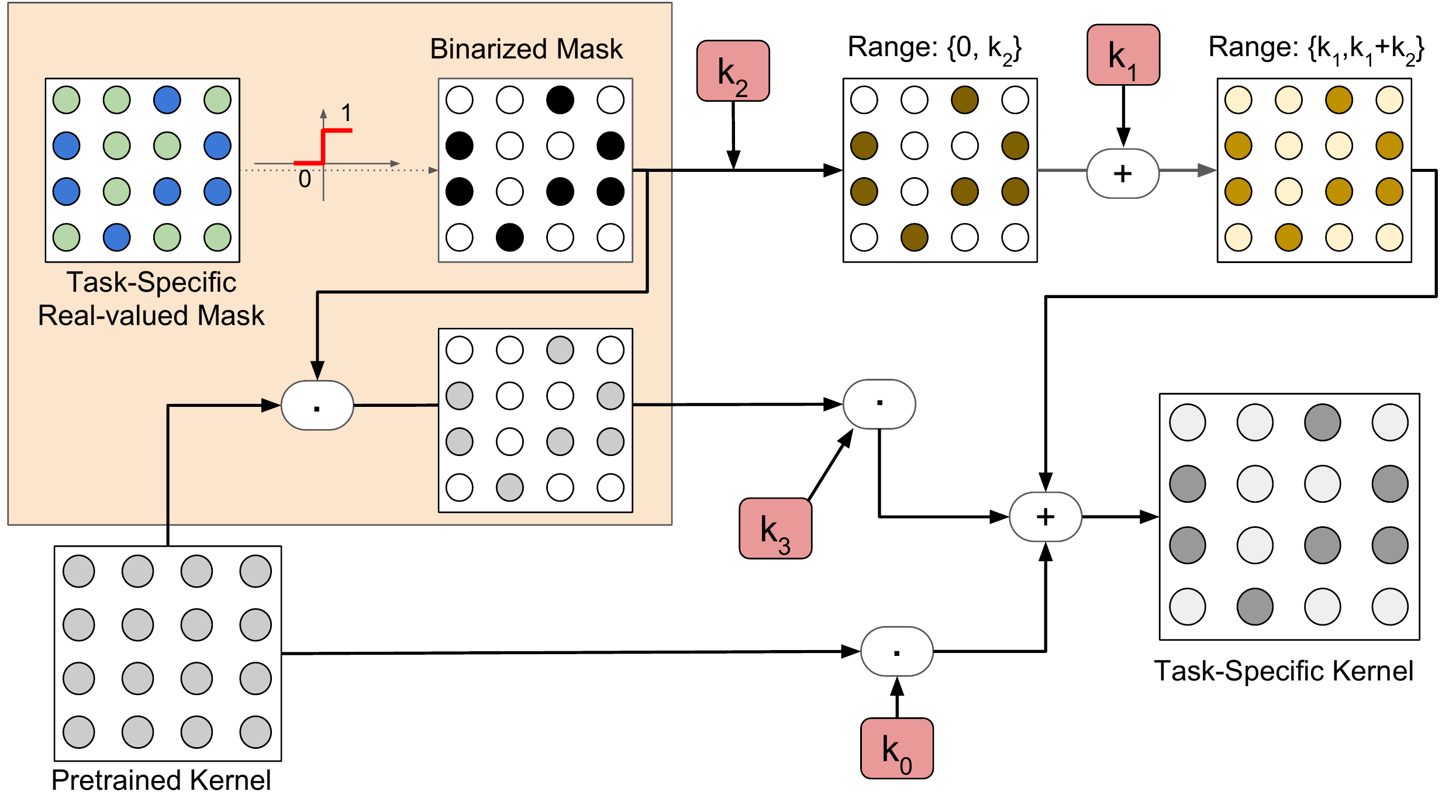}
     \caption{Overview of the proposed model (best viewed in color). Given a convolutional kernel, for each task, we exploit a real-valued mask to generate a task-specific binary mask. In previous works~\cite{mallya2018piggyback}, the binary mask is just used as a filter for the pretrained weights (orange block). We generalize this approach by including an affine transformation directly applied to the binary masks, which changes their range (through a scale parameter $k_2$) and their minimum value (through $k_1$). The original mask of \cite{mallya2018piggyback} and the pretrained kernel are scaled by the factors $k_3$ and $k_0$ respectively. All the different masks are summed to produce the final task-specific kernel. }
    \label{fig:method}
 \end{figure}

Our paper takes inspiration from this last work, upon which 
we build.
The idea is to formulate incremental learning as the problem of learning a perturbation of a \emph{baseline}, pre-trained network, in a way to maximize the performance on a new task. Moreover, the perturbation should be compact in the sense of limiting the number of additional parameters required with respect to the baseline network.
As opposed to~\cite{mallya2018piggyback}, where the authors perturb the baseline network by filtering the convolutional weights with a learned binary mask, we consider a more elaborated 
setting sketched in Figure~\ref{fig:method}, which actually generalizes the approach of~\cite{mallya2018piggyback}. We apply an affine transformation to each convolutional weight of the baseline network, which involves both a learned binary mask and few additional parameters. The binary mask in our case is not only used to filter the convolutional weights as in~\cite{mallya2018piggyback} (this becomes just one term in our transformation), but it is used also as a scaled and shifted additive component.
Our solution allows us to achieve two main goals: 1) boosting the performance of each task-specific network that we train, by leveraging the higher degree of freedom in perturbing the baseline network, while 2) keeping a low per-task overhead in terms of additional parameters (slightly more than 1 bit per parameter per task), mostly thanks to a better exploitation of the learned binary masks.

We assess the validity of our method, and some variants thereof, on standard benchmarks including the Visual Decathlon Challenge~\cite{RebuffiKSL17}. The experimental results show that our model achieves performances comparable with fine-tuning separate networks for each task on all benchmarks, while retaining a very small overhead in terms of additional parameters per task. Notably, we establish a new state-of-the-art result on the Visual Decathlon Challenge.



\section{Related works}\label{sec:related}
The keen interest on incremental and life-long learning methods dates back to the pre-convnet era, with shallow learning approaches ranging from large margin classifiers \cite{KuzborskijOC13,KuzborskijOC17} to non-parametric methods \cite{MensinkVPC13,RistinGGG16}. 
In recent years, various works have addressed the problem of incremental  and life-long learning  within the framework of deep architectures \cite{RebuffiKSL17,GuerrieroCM18,BendaleB16}. A major risk when training a neural network on a novel task is to deteriorate the performances of the network on old tasks, discarding previous knowledge. This phenomenon is called \textit{catastrophic forgetting} \cite{mccloskey1989catastrophic,french1999catastrophic,goodfellow2013empirical}. To alleviate this issue, various works designed constrained optimization procedures taking into account the initial network weights, trained on previous tasks. In \cite{li2017learning}, the authors exploit knowledge distillation \cite{hinton2015distilling} to obtain target objectives for previous tasks, while training for novel ones. The additional objective ensures preservation of the activation for previous tasks, making the model less prone to experience the catastrophic forgetting problem. In \cite{kirkpatrick2017overcoming} the authors consider to compute the update of the network parameters, based on their importance for previously seen tasks. 

While these approaches are optimal in terms of the required parameters, i.e. they maintain the same number of parameters of the original network, they limit the catastrophic forgetting problem to the expenses of a lower performance on both old and new tasks. Recent methods overcome this issue by devising task specific parameters which are added as new tasks are learned. If the initial network parameters remain untouched, the catastrophic forgetting problem is avoided but at the cost of the additional parameters required. The extreme case is the work of \cite{rusu2016progressive} in the context of reinforcement learning, where a parallel network is added each time a new task is presented with side task connections, exploited to improve the performances on novel tasks. 
In \cite{rebuffi2017learning}, task-specific residual components are added in standard residual blocks. In \cite{rosenfeld2017incremental} the authors propose to use controller modules where the parameters of the base architecture are recombined channel-wise. 

To provide more compact and less demanding overhead for the starting architecture, in \cite{mallya2017packnet} only a subset of network parameters is considered for each task. The intersection of the parameters used by different tasks is empty, thus the network can be trained end-to-end for each task. Obviously as the number of task increases, less parameters are available for each task, with a consequent limit on the performances of the network. Because of this, the authors proposed recently a more compact and effective solution \cite{mallya2018piggyback} with separate binary masks learned for each novel task. The binary masks determine which of the network parameters are useful for the new task and which are not. We take inspiration from this last work, but differently from \cite{mallya2018piggyback} we do not consider multiplicative binary masks. Our formulation, which generalizes~\cite{mallya2018piggyback}, allows us to use a comparable number of parameters per task with increased flexibility, due to the multiple ways in which the binary masks can be merged to the main architecture. This leads to improvements in performances, reducing the gap with the individual end-to-end trained architectures, while maintaining a low footprint per task in terms of number of additional parameters.

Due to the low overhead required by the task-specific parameters introduced by our method, our work is linked to recent works on binarization \cite{courbariaux2016binarized,hubara2016binarized,zhou2016dorefa,rastegari2016xnor,lin2015neural} and quantization \cite{hubara2016quantized,lin2016fixed} of network parameters. Opposite to these works, we do not compress the whole architecture or its activations but a subset of its parameters, those we add for each task. However, the optimization techniques used in these works (\eg \cite{hubara2016binarized}) are fundamental building blocks of our algorithm.

\section{Method}\label{sec:method}
We address the problem of incremental
learning 
of new tasks similarly to~\cite{mallya2018piggyback}, \ie we modify a \emph{baseline} network such as, \eg ResNet-50 pretrained on the ImageNet classification task, so to maximize its performance on a new task, while limiting the amount of additional parameters needed. The solution we propose exploits the key idea from Piggyback~\cite{mallya2018piggyback} of learning task-specific masks, but instead of pursuing the simple multiplicative transformation of the parameters of the baseline network, we define a parametrized, affine transformation mixing a binary mask and real parameters
that significantly increases the expressiveness of the approach, leading to a rich and nuanced ability to adapt the old parameters to the needs of the new tasks. This in turn brings considerable improvements on all the conducted experiments, as we will show in the experimental section, while retaining a reduced, per-task overhead.

\subsection{Overview}
Let us assume to be given a pre-trained, \emph{baseline} network $f_0(\cdot; \Theta, \Omega_0):\set X\to\set Y_0$ assigning a class label in $\set Y_0$ to elements of an input space $\set X$ (\eg images).\footnote{We focus on classification tasks, but the proposed method applies also to other tasks.} The parameters of the baseline network are partitioned into two sets: $\Theta$ comprises parameters that will be shared for other tasks, whereas $\Omega_0$ entails the rest of the parameters (\eg the classifier).
Our goal is to learn for each task $i\in\{1,\ldots,\con m\}$, with a possibly different output space $\set Y_i$,
a classifier $f_i(\cdot;\Theta,\Omega_i):\set X\to\set Y_i$. Here, $\Omega_i$ entails the parameters specific for the $i$th task, while $\Theta$ holds the shareable parameters of the baseline network mentioned above.
Before delving into the details of our method, we review the Piggyback solution presented in~\cite{mallya2018piggyback}.

 \begin{figure}[t]
 \centering
 \includegraphics[width=0.56\columnwidth]{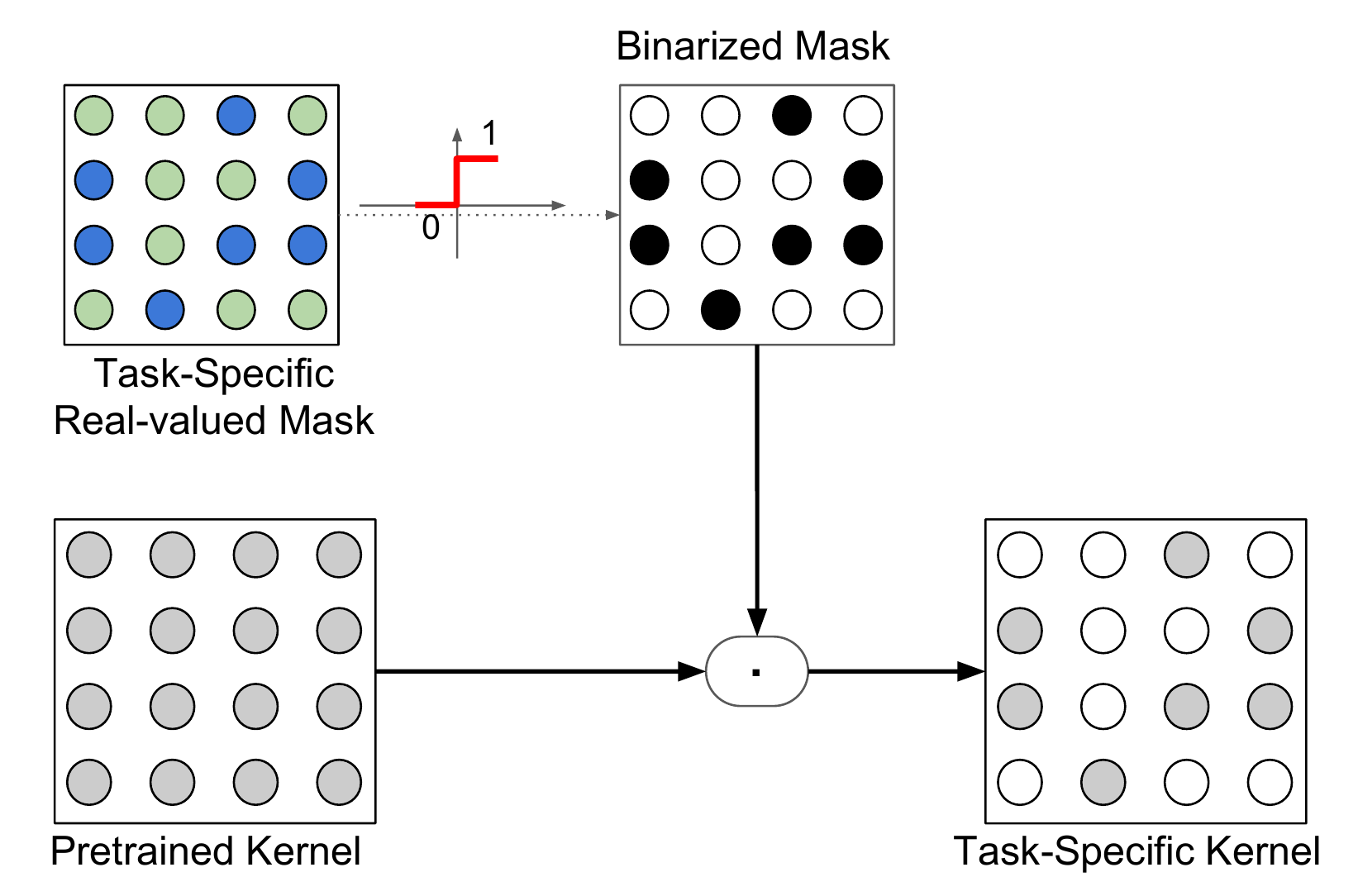}
     \caption{Overview of Piggyback \cite{mallya2018piggyback}. The binary mask is used as a filter for the pretrained weights, determining which weights of the original model are used for the novel task.} 
    \label{fig:piggyback}
 \end{figure}
\myparagraph{Piggyback~\cite{mallya2018piggyback}.} 
Each task-specific network $f_i$ shares the same structure of the baseline network $f_0$, except for having a possibly, differently sized classification layer.
All parameters of $f_0$, excepting the classifier, are shared across all the tasks. For each convolutional layer\footnote{Fully-connected layers are a special case.} of $f_0$ with parameters $\mat W$, the task-specific network $f_i$ holds a binary mask $\mat M$ that is used to mask $\mat W$ obtaining 
\begin{equation}\label{eq:piggy}
\hat{\mat W}=\mat W\circ \mat M\,,
\end{equation}
where $\circ$ is the Hadamard (or element-wise) product. 
The transformed parameters $\hat{\mat W}$ are then used in the convolutional layer of $f_i$. 
By doing so, the task-specific parameters that are stored in $\Omega_i$ amount to just a single bit per parameter in each convolutional layer, yielding a low overhead per additional task, while retaining a sufficient degree of freedom to build new convolutional weights. Figure \ref{fig:piggyback} gives an overview of the transformation.

\myparagraph{Proposed (full).}
The method we propose can be regarded as a parametrized generalization of Piggyback.
Similarly to~\cite{mallya2018piggyback}, we consider task-specific networks $f_i$ that are shaped as the baseline network $f_0$ and we store in $\Omega_i$ a binary mask $\mat M$ for each convolutional kernel $\mat W$ in the shared set $\Theta$. However, we depart from the simple multiplicative transformation of $\mat W$ used in \eqref{eq:piggy}, and consider instead a more general affine transformation of the base convolutional kernel $\mat W$ that depends on a binary mask $\mat M$ as well as additional parameters. Specifically, we transform $\mat W$ into 
\begin{equation}\label{eq:ours}
\tilde {\mat W}=k_0\mat W+k_1 \mathtt 1+k_2\mat M +k_3\mat W\circ\mat M\,,
\end{equation}
where $k_j\in\mathbb R$ are additional task-specific parameters in $\Omega_i$ that we learn along with the binary mask $\mat M$, and $\mat 1$ is an opportunely sized tensor of $1$s. 
We recover Piggyback if we set $k_{0,1,2}=0$ and $k_3=1$. Figure~\ref{fig:method} provides an overview of the transformation in \eqref{eq:ours}.

 \begin{figure}[t]
 \centering
 \includegraphics[width=0.8\columnwidth]{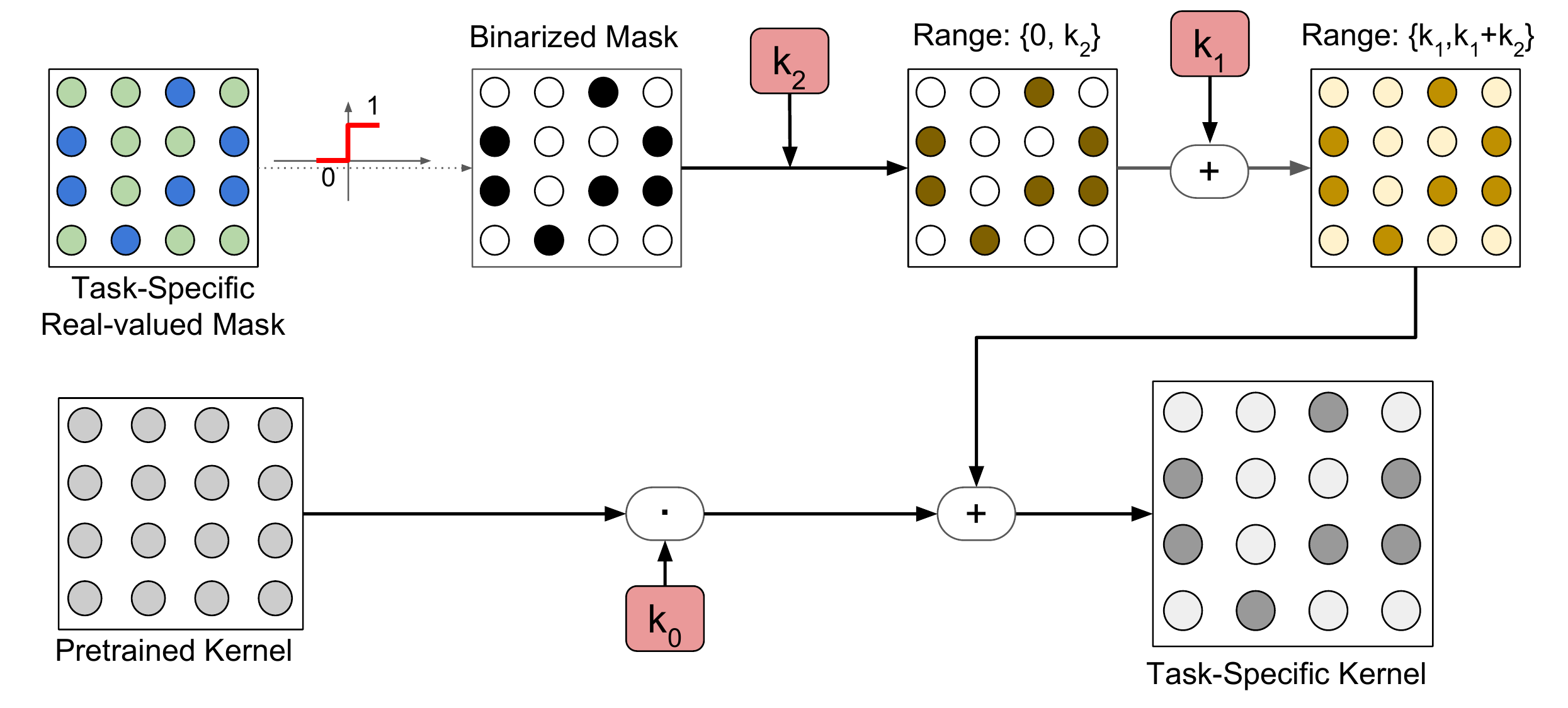}
     \caption{Simple version of our model. An affine transformation scale and translate the binary masks through the parameters $k_2$ and $k_1$ respectively. The obtained mask is summed to the pretrained kernel in order to obtain the final task-specific weights.} 
    \label{fig:simple}
 \end{figure}
\myparagraph{Proposed (simple).} The transformation in \eqref{eq:ours} includes a term that corresponds to the Piggyback's multiplicative transformation $\hat {\mat W}$ scaled by $k_3$. However, it turns out that empirically this term is not essential to achieve state-of-the-art performance as we will show in the experimental section. For this reason we put special focus on a simplified version of \eqref{eq:ours} that gets rid of the Piggyback component, \ie we set $k_3=0$.
This yields the following simplified transformation
\begin{equation}\label{eq:simple}
\check {\mat W}=k_0\mat W+k_1 \mathtt 1+k_2\mat M\,,
\end{equation}
which will be taken into account in our analysis. Figure \ref{fig:simple} illustrates the transformation above.

\vspace{10pt}
Besides learning the binary masks and the parameters $k_j$, unless otherwise stated, we opt also for task-specific batch-normalization parameters (\ie mean, variance, scale and bias), which will be part of $\Omega_i$, and thus optimized for each task, rather than being fixed in $\Theta$. 
In the cases where we have a convolutional layer followed by batch normalization, we keep the corresponding parameter $k_0$ fixed to $1$, because the output of batch normalization is invariant to the scale of the convolutional weights.

The additional parameters that we introduce with our method bring a negligible per-task overhead compared to Piggyback, which is nevertheless generously balanced out by a significant boost of the performance of the task-specific classifiers.

\subsection{Learning Binary Masks}
We learn the parameters $\Omega_i$ of each task-specific network $f_i$ by minimizing the classification log-loss, given a training set, using standard, stochastic optimization methods (more details are given in Section~\ref{sec:experiments}). However, special care should be taken for the optimization of the binary masks.

Instead of optimizing the binary masks directly, which would turn the learning into a combinatorial problem, we apply the solution adopted in~\cite{mallya2018piggyback}, \ie we replace each binary mask $\mat M$ with a thresholded real matrix $\mat R$. By doing so, we shift from optimizing discrete variables in $\mat M$ to continuous ones in $\mat R$. However, the gradient of the hard threshold function $h(r)=1_{r\geq 0}$ is zero almost everywhere, which makes this solution apparently incompatible with gradient-based optimization approaches. To sidestep this issue we consider a strictly increasing, surrogate function $\tilde h$ that will be used in place of $h$ \emph{only} for the gradient computation, \ie \[
h'(r)\approx \tilde h'(r)\,,
\]
where $h'$ denotes the derivative of $h$ with respect to its argument.
The gradient that we obtain via the surrogate function has the property that it always points in the right down hill direction in the error surface. To see this, let's focus on a single entry $r$ of $\mat R$, let $m=h(r)$ and let $E(m)$ be the error function. Then 
\[
\sign((E\circ h)'(r))=\sign(E'(m)h'(r))=\sign\left(E'(m)\tilde h'(r)\right)
\]
and, since $\tilde h'(r)>0$ by construction of $\tilde h$, we obtain the sign agreement
\[
\sign\left((E\circ h)'(r)\right)=\sign\left(E'(m)\right)\,.
\]
Accordingly, when the gradient of $E(h(r))$ with respect to $r$ is positive (negative), this induces a decrease (increase) of $r$. By the monotonicity of $h$ this eventually induces a decrease (increase) of $m$, which is compatible with the direction pointed by the gradient of $E$ with respect to $m$.

By taking $\tilde h(x)=x$, \ie the identity function, we recover the workaround suggested in~\cite{hin12}, and employed also in~\cite{mallya2018piggyback}. By taking $\tilde h(x)=(1+e^{-x})^{-1}$, \ie the sigmoid function, we obtain a better approximation that has been suggested in~\cite{goodman1994learning,bengio2013estimating}.

\section{Experiments}\label{sec:experiments}
\vspace{-10pt}
\myparagraph{Datasets.}
In the following we test our method on two different benchmarks. For the first benchmark we follow \cite{mallya2018piggyback}, and we use 6 datasets: ImageNet \cite{russakovsky2015imagenet}, VGG-Flowers \cite{nilsback2008automated}, Stanford Cars \cite{krause20133d}, Caltech-UCSD Birds (CUBS) \cite{wah2011caltech}, Sketches \cite{eitz2012humans} and WikiArt \cite{saleh2015large}. VGG-Flowers \cite{nilsback2008automated} is a dataset of fine-grained recognition containing images of 102  categories, corresponding to different kind of flowers. There are 2'040 images for training and 6'149 for testing. Stanford Cars \cite{krause20133d} contains images of 196 different types of cars with approximately 8 thousand images for training and 8 thousands for testing. Caltech-UCSD Birds \cite{wah2011caltech} is another dataset of fine-grained recognition containing images of 200 different species of birds, with approximately 6 thousands images for training and 6 thousands for testing. Sketches \cite{eitz2012humans} is a dataset composed of 20 thousands sketch drawings, 16 thousands for training and 4 thousands for testing. It contains images of 250 different objects in their sketched representations. WikiArt \cite{saleh2015large} contains painting from 195 different artists. The dataset has 42129 images for training and 10628 images for testing. These datasets contain a lot of variations both from the category addressed (\ie cars \cite{krause20133d} vs birds \cite{wah2011caltech}) and the appearance of their instances (from natural images \cite{russakovsky2015imagenet} to paintings \cite{saleh2015large} and sketches \cite{eitz2012humans}), thus representing a challenging benchmark for incremental learning techniques. 

The second benchmark 
is the Visual Decathlon Challenge \cite{rebuffi2017learning}. This challenge has been introduced in order to check the capability of a single algorithm to tackle 10 different classification tasks. The tasks are taken from the following datasets: ImageNet \cite{russakovsky2015imagenet}, CIFAR-100 \cite{krizhevsky2009learning}, Aircraft \cite{maji2013fine}, Daimler pedestrian classification (DPed) \cite{munder2006experimental}, Describable textures (DTD) \cite{cimpoi2014describing}, German traffic signs (GTSR) \cite{stallkamp2012man} , Omniglot \cite{lake2015human}, SVHN \cite{netzer2011reading}, UCF101 Dynamic Images \cite{bilen2016dynamic,soomro2012ucf101} and VGG-Flowers \cite{nilsback2008automated}. A more detailed description of the challenge and the datasets can be found in \cite{rebuffi2017learning}. For this challenge, an independent scoring function is defined \cite{rebuffi2017learning}. This function $S$ is expressed as:
\begin{equation}\label{eq:S}
S=\sum^{10}_{d=1}\alpha_d\text{max}\{0,E_d^{\text{max}}-E_d\}^{2}
\end{equation}
where $E_d^{\text{max}}$ is the test error of the baseline in the domain $d$, $E_d$ is the test error of the submitted model and $\alpha$ is a scaling parameter ensuring that the perfect score for each task is 1000, thus with a maximum score of 10000 for the whole challenge. The baseline error is computed doubling the error of 10 independent models fine-tuned on the single tasks.
This score function takes into account the performances of a model on all 10 classes, preferring models with good performances on all of them compared to models outperforming by a large margin the baseline in just few of them. 

\myparagraph{Networks and training protocols.} For the first benchmark, we use 2 networks: ResNet-50 \cite{he2016deep} and DenseNet-121 \cite{huang2017densely}, and we report the results of Piggyback \cite{mallya2018piggyback}, PackNet \cite{mallya2017packnet} and both the simple and full version of our model described in Section~\ref{sec:method}. Following the protocol of \cite{mallya2018piggyback}, for all the models we start from the networks pre-trained on ImageNet and train the task-specific networks using Adam \cite{kingma2014adam} as optimizer except for the classifiers where SGD \cite{bottou2010large} with momentum is used. The networks are trained with a batch-size of 32 and an initial learning rate of 0.0001 for Adam and 0.001 for SGD with momentum 0.9. Both the learning rates are decayed by a factor of 10 after 15 epochs. In this scenario we use input images of size $224\times224$ pixels, with the same data augmentation (\ie mirroring and random rescaling) of \cite{mallya2017packnet,mallya2018piggyback}. The real-valued masks are initialized with random values drawn from a uniform distribution with values between $0.0001$ and $0.0002$. Since our model is independent on the order of the tasks, we do not take into account different possible orders, reporting the results as accuracy averaged across multiple runs. For simplicity, in the following we will denote this scenario as \textit{ImageNet-to-Sketch}.

For the Visual Decathlon we employ the Wide ResNet-28 \cite{zagoruyko2016wide} adopted by previous methods \cite{rebuffi2017learning,rosenfeld2017incremental,mallya2018piggyback}, with a widening factor of 4 (\ie 64, 128 and 256 channels in each residual block). Following \cite{rebuffi2017learning} we rescale the input images to $72\times72$ pixels giving as input to the network images cropped to $64\times64$. We follow the protocol in \cite{mallya2018piggyback}, by training the simple and full versions of our model for 60 epochs with a batch-size of 32, and using again Adam for the entire architecture but the classifier, where SGD with momentum is used. The same learning rates of the first benchmark are adopted and are decayed by a factor of 10 after 45 epochs. The same initialization scheme is used for the real-valued masks. \textit{No hyperparameter tuning} has been performed as we used a single training schedule for all the 10 tasks, except for the ImageNet pretrained model, which was trained following the schedule of \cite{rebuffi2017learning}. As for data augmentation, mirroring has been performed, except for the datasets with digits (\ie SVHN), signs (Omniglot, GTSRB) and textures (\ie DTD) as it may be rather harmful (as in the first 2 cases) or not necessary.

\subsection{Results}
\vspace{-10pt}
\myparagraph{ImageNet-to-Sketch.} In the following we discuss the results obtained by our model on the ImageNet-to-Sketch scenario. We compare our method with Piggyback \cite{mallya2018piggyback}, PackNet \cite{mallya2017packnet} and two baselines considering the network only as feature extractor, training only the task-specific classifier, and individual networks separately fine-tuned on each task. PackNet \cite{mallya2017packnet} adds a new task to an architecture by identifying important weights for the task, optimizing the architecture through alternated pruning and re-training steps. Since this algorithm is dependent on the order of the task, we report the performances for two different orderings \cite{mallya2018piggyback}: starting from the model pre-trained on ImageNet, in the first setting ($\downarrow$) the order is CUBS-Cars-Flowers-WikiArt-Sketch while for the second ($\uparrow$) the order is reversed. For our model, we evaluate both the full and the simple version, including task-specific batch-normalization layers. Since including batch-normalizatin layers affects the performances, for the sake of presenting a fair comparison, we report also the results of Piggyback \cite{mallya2018piggyback} obtained as a special case of our model with separate BN parameters per task.

Results are reported in Tables \ref{tab:resnet-ImageNet} and \ref{tab:densenet-ImageNet}. We see that both versions of our model are able to fill the gap between the classifier only baseline and the individual fine-tuned architectures, almost entirely in all settings. For larger and more diverse datasets such as Sketch and WikiArt, the gap is not completely covered, but the distance between our models and the individual architectures is always less than 1\%. These results are remarkable given the simplicity of our method, not involving any assumption of the optimal weights per task \cite{mallya2017packnet,li2017learning}, and the small overhead in terms of parameters that we report in the row "\#~Params" (\ie $1.17$ for ResNet-50 and $1.21$ for DenseNet-121), which represents the total number of parameters (counting all tasks and excluding the classifiers) relative to the ones in the baseline network.
For what concerns the comparison with the other algorithms, our model consistently outperforms both the basic version of Piggyback and PackNet in all the settings and architectures, with the exception of Sketch for the DenseNet architecture, in which the performances are comparable with those of Piggyback. When task-specific BN parameters are introduced also for Piggyback, the gap in performances is reduced, with comparable performances in some settings (\ie CUBS) but with still large gaps in others (\ie Flowers, Stanford Cars and WikiArt). These results show that the advantages of our model are not only due to the additional BN parameters, but also to the more flexible and powerful affine transformation introduced. 

Both Piggyback and our model outperform PackNet and, as opposed to the latter method, do not suffer from the heavily dependence on the ordering of the tasks. This advantage stems from having an incremental learning strategy that is task independent, with the base network not affected by the new tasks that are learned.
             

\begin{table}[t]
			\caption{Accuracy of ResNet-50 architectures in the ImageNet-to-Sketch scenario.} 
                   
		\centering
        \scalebox{1}{
		\begin{tabular}{ l | c || ec | ec | ec | ec | ec | ec || c  } 
			\hline
			\multirow{2}{*}{Dataset} & Classifier & \multicolumn{2}{ c | }{PackNet\cite{mallya2018piggyback}} &\multicolumn{2}{c |}{Piggyback} & \multicolumn{2}{ c ||}{Ours} & Individual\\
          & Only \cite{mallya2018piggyback} &$\downarrow$&$\uparrow$ & \cite{mallya2018piggyback}&BN& Simple& Full  & \cite{mallya2018piggyback}\\ \hline
       ImageNet &76.2&75.7&75.7&\textbf{76.2} &\textbf{76.2}&\textbf{76.2}&\textbf{76.2} &76.2\\
       CUBS&70.7 &80.4&71.4&80.4 &82.1&\textbf{82.6} &82.4&82.8 \\
       Stanford Cars&52.8&86.1&80.0 &88.1 &90.6&\textbf{91.5}&91.4&91.8 \\
       Flowers&86.0&93.0&90.6&93.5&95.2&96.5&\textbf{96.7} &96.6\\
       WikiArt&55.6&69.4&70.3&73.4 &74.1&74.8&\textbf{75.3}&75.6\\
       Sketch&50.9&76.2&78.7&79.4 &79.4&\textbf{80.2}&\textbf{80.2} &80.8 \\
            \hline
             \# Params&1&\multicolumn{2}{c |}{1.10}&1.16&1.17& 1.17&1.17&6\\  \hline
		\end{tabular}
        }
		\label{tab:resnet-ImageNet}
\end{table}
\begin{table}[t]
			\caption{Accuracy of DenseNet-121 architectures in the ImageNet-to-Sketch scenario.} 
\centering
		\begin{tabular}{ l | c || ec | ec| ec | ec | ec | ec || c  } 
			\hline
			\multirow{2}{*}{Dataset} & Classifier &  \multicolumn{2}{c |}{PackNet\cite{mallya2018piggyback}}&\multicolumn{2}{c |}{Piggyback} & \multicolumn{2}{c ||}{Ours} & Individual\\
          & Only \cite{mallya2018piggyback}&$\downarrow$ & $\uparrow$&\cite{mallya2018piggyback} &BN&Simple&Full & \cite{mallya2018piggyback}\\ \hline
       ImageNet &74.4 &\textbf{74.4}&\textbf{74.4}&\textbf{74.4} &\textbf{74.4}&\textbf{74.4} & \textbf{74.4}&74.4\\
       CUBS &73.5 &80.7&69.6&79.7 &81.4&81.5&\textbf{81.7}&81.9\\
       Stanford Cars&56.8 &84.7&77.9&87.2 &90.1&\textbf{91.7} &91.6&91.4\\
       Flowers&83.4 &91.1&91.5&94.3 &95.5&96.7&\textbf{96.9} &96.5\\
       WikiArt&54.9 &66.3&69.2 &72.0&73.9&75.5&\textbf{75.7} &76.4\\
       Sketch&53.1 &74.7&78.9&\textbf{80.0} &79.1&79.9&79.8 &80.5 \\
            \hline
           \# Params&1&\multicolumn{2}{c |}{1.11}&1.15& 1.21&1.21&1.21&6 \\  \hline
		\end{tabular}
		\label{tab:densenet-ImageNet}
        \vspace{-5pt}
\end{table}

\myparagraph{Visual Decathlon Challenge.}
In this section we report the results obtained on the Visual Decathlon Challenge. We compare our model with the baseline method Piggyback \cite{mallya2018piggyback} (PB),  the improved version of the winner entry of the 2017 edition of the challenge \cite{rosenfeld2017incremental} (DAN) and the baselines proposed by the authors of the challenge \cite{rebuffi2017learning}. For the latter, we report the results of 5 models: the network used as feature extractor (Feature), 10 different models fine-tuned on each single task (Finetune), the network with task-specific residual adapter modules \cite{rebuffi2017learning} (RA), the same model with increased weight decay (RA-decay) and the same architecture jointly trained on all 10 tasks, in a round-robin fashion (RA-joint). The first two models are considered as references. Similarly to \cite{rosenfeld2017incremental} we tune the training schedule, jointly for the 10 tasks, using the validation set, and evaluate the results obtained on the test set (via the challenge evaluation server) by a model trained on the union of the training and validation sets, using the validated schedule. As opposed to methods like~\cite{rebuffi2017learning} we use the same schedule for the 9 tasks (except for the baseline pretrained on ImageNet), without adopting task-specific strategies for setting the hyper-parameters. Moreover, we do not employ our algorithm while pretraining the ImageNet architecture as in~\cite{rebuffi2017learning}. For fairness, we additionally report the results obtained by our implementation of \cite{mallya2018piggyback} using the same pretrained model, training schedule and data augmentation adopted for our algorithm (PB ours).

The results are reported in Tables \ref{tab:vdc-scores} and \ref{tab:vdc-accuracy} in terms of the $S$-score (see, \eqref{eq:S}) and accuracy, respectively. From the first table we can see that our model establishes a new state-of-the-art for the Visual Decathlon competition both in its simple (S) and full (F) form, with a gain of more than 400 points on the previous state-of-the-art-model \cite{rosenfeld2017incremental} for the simple version and of more than 600 points for the full version. Looking at the partial results, excluding the ImageNet baseline, our model achieves the best scores in 3 out of 9 tasks (Cifar-100, GTSR, VFF-Flowers), with the full model achieving either the top-2 or comparable results in 7 out of 9 tasks. The only exceptions are UCF-101, where  we achieve slightly lower results and Aircraf, where our models suffer a high accuracy drop. In this last case, by looking at Table \ref{tab:vdc-accuracy} it is possible to notice a drop of almost 13 points in accuracies with respect to the best model \cite{mallya2018piggyback}. Tuning the hyperparameters (\eg weight-decay which we do not use) could cover this gap, but this is out of the scope of this work. Despite this drop, our model achieves an average accuracy over the 10 tasks higher than previous incremental learning strategies, and on par with the model fine-tuned on all tasks together (RA-joint). Interestingly, while our simple model achieves comparable average accuracy with respect to other models (\eg PB, DAN, RA-decay), it obtains a much higher decathlon score. This highlights its capabilities of tackling all 10 tasks with good results, without peaked accuracies on just few of them.

\begin{table}[t]
			\caption{Results in terms of decathlon score for the Visual Decathlon Challenge. Best model in bold, second best underlined.} 
            
		\centering
		\scalebox{.9}{\begin{tabular}{ l | c || c  c  c  c  c  c  c  c  c  c || c   } 
			\hline Method&Params&ImNet&Airc. &C100&DPed&DTD&GTSR&Flwr.&Oglt.&SVHN&UCF&Score\\\hline
            Feature\cite{rebuffi2017learning}&1&247&1&0&0&149&0&85&0&0&62&544\\
            Finetune\cite{rebuffi2017learning}&10&250&250&250&250&250&250&250&250&250&250&2500\\
            \hline
            RA\cite{rebuffi2017learning}&2&247&206&225&329&200&163&8&335&192&213&2118\\
            RA-decay\cite{rebuffi2017learning}&2&247&270&225&330&268&258&257&335&192&\underline{239}&2621\\
            RA-joint\cite{rebuffi2017learning}&2&242&295&228&285&267&237&307&\textbf{344}&197&\textbf{241}&2643\\
            DAN \cite{rosenfeld2017incremental}&2.17&224	&\underline{300}	&196	&155	&262	&473	&390	&\underline{337}	&283	&232	&2852\\
            PB\cite{mallya2018piggyback}&1.28&224	&\textbf{316}	&191	&\textbf{624}	&272	&200	&191	&247	&\textbf{359}	&213	&2838\\
            PB ours&1.28&\textbf{262}&159&195&430&298&533&\textbf{292}&155&273&208&2805\\
          	Ours (S)&1.29&\textbf{262}	&149	&\underline{244}	&401	&\underline{291}	&\underline{665}	&\underline{459}	&320	&247	&225	&\underline{3263}\\
            Ours (F)&1.29&\textbf{262}	&164	&\textbf{246}	&\underline{535}	&287	&\textbf{708}	&\textbf{466}	&314	&\underline{291}	&224	&\textbf{3497}\\
            \hline
             
		\end{tabular}}
		\label{tab:vdc-scores}
                \vspace{-10pt}
\end{table}

\begin{table}[t]
			\caption{Results in terms of accuracy for the Visual Decathlon Challenge. Best model in bold, second best underlined.} 
            
		\centering
        \scalebox{.9}{
		\begin{tabular}{ l | c || c  c  c  c  c  c  c  c  c  c || c } 
			\hline Method&Params&ImNet&Airc. &C100&DPed&DTD&GTSR&Flwr.&Oglt.&SVHN&UCF&Mean\\\hline
            Feature \cite{rebuffi2017learning}&1&59.7	&23.3	&63.1	&80.3	&45.4	&68.2	&73.7	&58.8	&43.5	&26.8	&54.3\\
            Finetune \cite{rebuffi2017learning}&10&59.9	&60.3	&82.1	&92.8	&55.5	&97.5	&81.4	&87.7	&96.6	&51.2	&76.5\\
      \hline     RA\cite{rebuffi2017learning}&2&59.7	&56.7	&81.2	&93.9	&50.9	&97.1	&66.2	&89.6	&96.1	&47.5	&73.9\\
           RA-decay\cite{rebuffi2017learning}&2&59.7	&61.9	&81.2	&93.9	&57.1	&97.6	&81.7	&89.6	&96.1	&\underline{50.1}	&76.9\\
           RA-joint\cite{rebuffi2017learning}&2&59.2	&63.7	&81.3	&93.3	&57.0	&97.5	&83.4	&\textbf{89.8}	&96.2	&\textbf{50.3}	&\textbf{77.2}\\
            DAN \cite{rosenfeld2017incremental}&2.17&57.7	&\underline{64.1}	&80.1	&91.3	&56.5	&98.5	&86.1	&\underline{89.7}	&96.8	&49.4	&77.0\\
            PB \cite{mallya2018piggyback}&1.28&57.7	&\textbf{65.3}	&79.9	&\textbf{97.0}	&57.5	&97.3	&79.1	&87.6	&\textbf{97.2}	&47.5	&76.6\\
            PB ours &1.28&\textbf{60.8}&52.3&80.0&95.1&\textbf{59.6}&98.7&82.9&85.1&96.7&46.9&75.8\\
          	Ours (S)&1.29&\textbf{60.8}	&51.3	&\underline{81.9}	&94.7	&\underline{59.0}	&\underline{99.1}	&\underline{88.0}	&89.3	&96.5	&48.7	&76.9\\
            Ours (F)&1.29&\textbf{60.8}	&52.8	&\textbf{82.0}	&\underline{96.2}	&58.7	&\textbf{99.2}	&\textbf{88.2}	&89.2	&\underline{96.8}	&48.6	&\textbf{77.2}\\
            \hline
             
		\end{tabular}}
		\label{tab:vdc-accuracy}
        \vspace{-10pt}
\end{table}
\subsection{Ablation Study}
\label{sec:ablation}

In the following we will analyze the impact of the various components of our model. In particular we consider the impact of the parameters $k_0$, $k_1$, $k_2$, $k_3$ and the surrogate function $\tilde{h}$ on the final results of our model for the ResNet-50 architecture in the ImageNet-to-Sketch scenario. Since the architecture contains batch-normalization layers, we set $k_0=1$ for our simple and full versions and $k_0=0$ when we analyze the special case \cite{mallya2018piggyback}. For the other parameters we adopt various choices: either we fix them to a constant in order not take into account their impact, or we train them, to assess their particular contribution to the model. The results of our analysis are shown in Table \ref{tab:ablation-full}. 

As the Table shows, while the BN parameters allow a boost in the performances of Piggyback, adding $k_1$ to the model does not provide further gain in performances. This does not happen for the simple version of our model: without $k_1$ our model is not able to fully exploit the presence of the binary masks, achieving even lower performances with respect to the Piggyback model. As $k_1$ is introduced, the boost of performances is significant such that neither the inclusion of $k_2$, nor considering channel-wise parameters $k_1$ provide further gains. Slightly better results are achieved in a larger datasets, such as WikiArt, with the additional parameters giving more capacity to the model, thus better handling the larger amount of information available in the dataset. 

As to what concerns the choice of the surrogate $\tilde{h}$, no particular advantage has been noted when $\tilde{h}(x)=\sigma(x)$ with respect to the standard straight-through estimator ($\tilde{h}(x)=x$). This may be caused by the noisy nature of the straight-through estimator, which has the positive effect of regularizing the parameters, as shown in previous works \cite{bengio2013estimating,neelakantan2015adding}.
\begin{table}[t]
			\caption{Impact of the parameters $k_0$, $k_1$. $k_2$ and $k_3$ of our model using the ResNet-50 architectures in the ImageNet-to-Sketch scenario. \ding{51} denotes a learned parameter, while~$^*$ denotes \cite{mallya2018piggyback} obtained as a special case of our model.} 

		\centering
		\begin{tabular}{ l | c | c | c | c | c | c | c | c | c |  } 
        Method &$k_0$ &$k_1$ & $k_2$&$k_3$& CUBS & CARS & Flowers & WikiArt & Sketch\\
         \hline
Piggyback \cite{mallya2018piggyback}&0&0&0&1&80.4&88.1&93.6&73.4&79.4\\
\hline
Piggyback$^*$ 
&0&0&0&1&80.4&87.8&93.1&72.5&78.6\\ 
Piggyback$^*$ with BN & 0&0&0&1 &82.1&90.6&95.2&74.1&79.4\\ 
Piggyback$^*$ with BN
&0&\ding{51}&0&1&81.9&89.9&94.8&73.7&79.9\\ 
        Ours (Simple, no bias)
&1&0&\ding{51}&0&80.8&90.3&96.1&73.5&80.0\\ 
        Ours (Simple) &1&\ding{51}&\ding{51}&0&82.6&91.5&96.5&74.8&80.2\\
        Ours (Simple with Sigmoid) &1&\ding{51}&\ding{51}&0&82.6&91.4&96.4&75.2&80.2\\
        Ours (Full)&1&\ding{51}&\ding{51}&\ding{51}&82.4&91.4&96.7&75.3&80.2\\
         Ours (Full with Sigmoid)&1&\ding{51}&\ding{51}&\ding{51}&82.7&91.4&96.6&75.2& 80.2\\
        Ours (Full, channel-wise) &1&\ding{51}&\ding{51}&\ding{51}&82.0&91.0&96.3&74.8&80.0\\
        
			\hline
             
		\end{tabular}
		\label{tab:ablation-full}
        \vspace{-10pt}
\end{table}

\subsection{Parameter Analysis}
\begin{figure}[!b]
\centering
 \includegraphics[width=1.\textwidth,trim=2cm 0 1.5cm 0,clip]{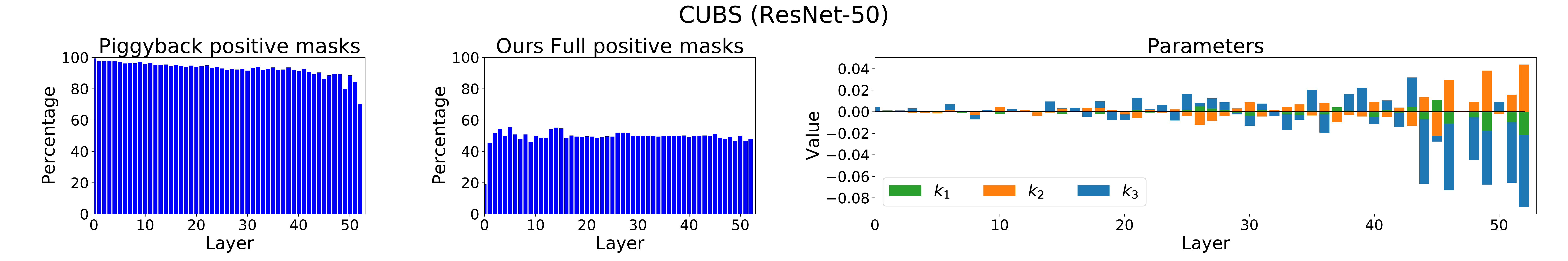}  \\
\includegraphics[width=1.\textwidth,trim=2cm 0 1.5cm 0,clip]{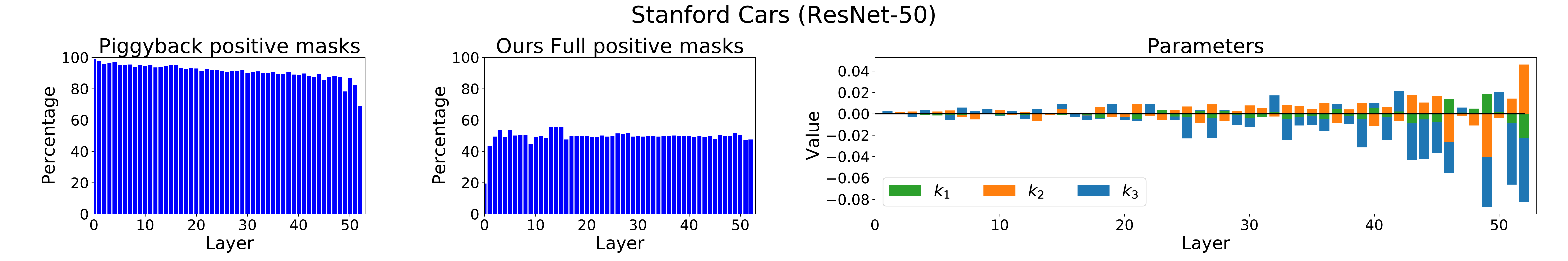} \\
\includegraphics[width=1.\textwidth,trim=2cm 0 1.5cm 0,clip]{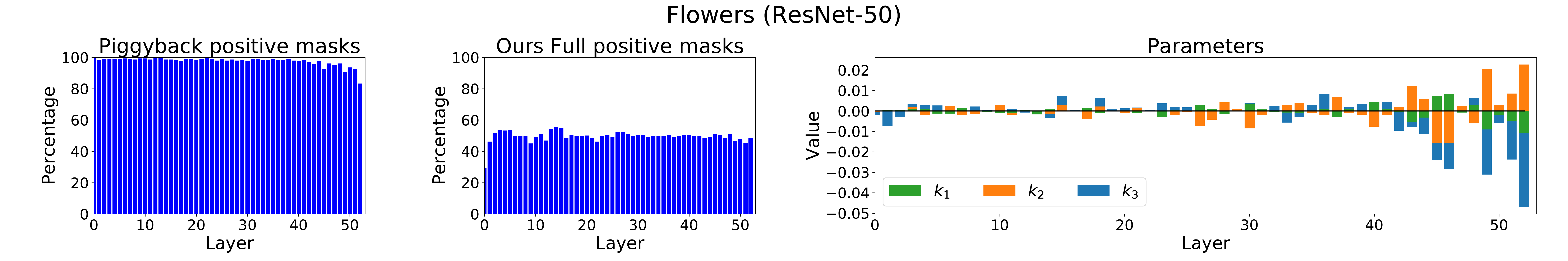} \\
   \includegraphics[width=1.\textwidth,trim=2cm 0 1.5cm 0,clip]{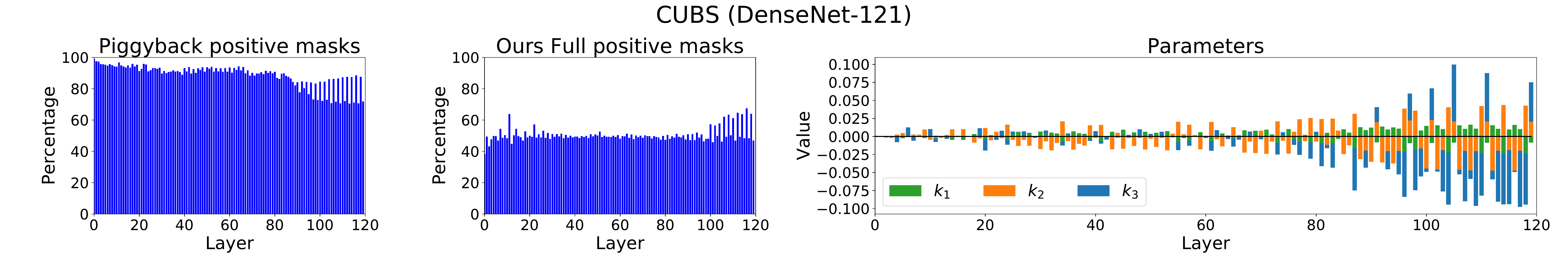}  \\
\includegraphics[width=1.\textwidth,trim=2cm 0 1.5cm 0,clip]{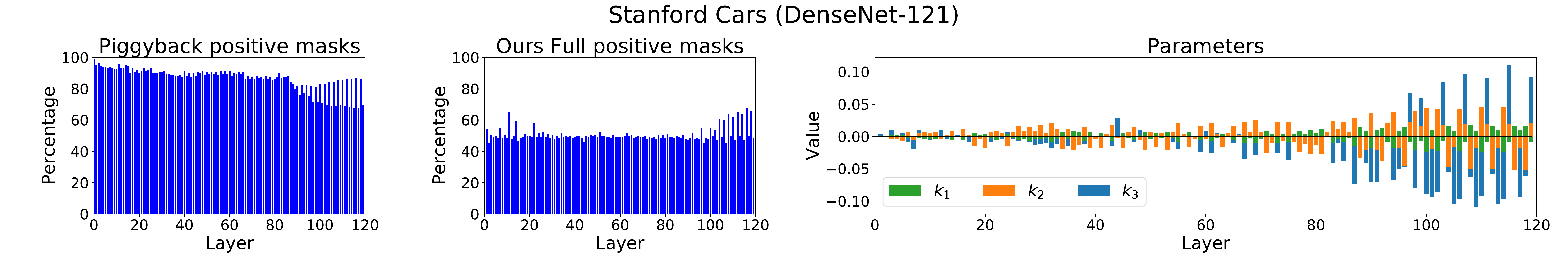} \\
\includegraphics[width=1.\textwidth,trim=2cm 0 1.5cm 0,clip]{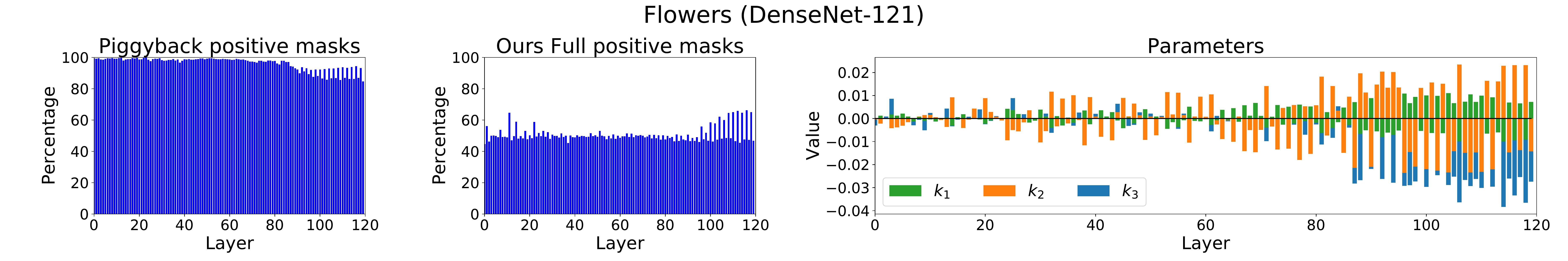} \\
  \caption{Percentage of 1s in the binary masks at different layers depth for Piggyback (left) and our full model (center) and values  of the parameters $k_1$, $k_2$, $k_3$ computed by our full model (right) for three datasets (CUBS, Standford Cars and Flowers) of the Imagenet-to-Sketch benchmark and the 2 architectures (ResNet-50 and DenseNet-121).
  }
  \label{fig:resnet-params}
\end{figure}

We analyze the values of the parameters $k_1$, $k_2$ and $k_3$ of one instance of our full model in the ImageNet-to-Sketch benchmark. We use both the architectures employed in that scenario (\ie ResNet-50 and DenseNet-121) and we plot the values of  $k_1$, $k_2$ and $k_3$ as well as the percentage of 1s present inside the binary masks for different layers of the architectures. Together with those values we report the percentage of 1s for the masks obtained through our implementation of Piggyback. Both the models have been trained considering task-specific batch-normalization parameters. The results are shown in Figure \ref{fig:resnet-params}. In all scenarios our model keeps almost half of the masks active across the whole architecture. Compared to the masks obtained by Piggyback, there are 2 differences: 1) Piggyback exhibits denser masks (\ie with a larger portion of 1s), 2) the density of the masks in Piggyback tends to decreases as the depth of the layer increases. Both these aspects may be linked to the nature of our model: by having more flexibility through the affine transformation adopted, there is less need to keep active large part of the network, since a loss of information can be recovered through the other components of the model, as well as constraining a particular part of the architecture. For what concerns the value of the parameters $k_1$, $k_2$ and $k_3$ for both architectures $k_2$ and $k_3$ tend to have larger magnitudes with respect to $k_1$. Also, the values of $k_2$ and $k_1$ tend to have a different sign, which allows the term $k_1\mat 1+k_2\mat M$ to span over positive and negative values. We also notice that the transformation of the weights are more prominent as the depth increases, which is intuitively explained by the fact that baseline network requires stronger adaptation to represent the higher-level concepts pertaining to different tasks.

\section{Conclusions}\label{sec:conclusions}
This work presents a simple yet powerful method for learning incrementally new tasks, given a fixed, pre-trained deep architecture. We build on the intuition of \cite{mallya2018piggyback}, and generalized the idea of masking the original weights of the network with learned binary masks. By introducing an affine transformation that acts upon such weights, we allow for a richer set of possible modifications of the original network that in turn makes it possible to capture better the characteristics of the new tasks. Experiments on two public benchmarks fully confirm the power of our approach. In particular, our algorithm reaches the new state of the art in the Visual Decathlon Challenge, with an advantage of more that 600 points compared to the previously reported best result. 

Future work will explore the possibility to exploit this approach on several life long learning scenarios, from incremental class learning to open world recognition. An interesting research direction would be exploring and exploiting the relationship between different task through cross-task affine transformations, in order to reuse previous knowledge collected through different tasks by the model.

\clearpage

\bibliographystyle{splncs03}
\bibliography{egbib}
\end{document}